\documentclass{article}
\usepackage{ijcai23}

\usepackage{times}
\usepackage{soul}
\usepackage{url}
\usepackage[hidelinks]{hyperref}
\usepackage[utf8]{inputenc}
\usepackage[small]{caption}
\usepackage{graphicx}
\usepackage{amsmath}
\usepackage{amsthm}
\usepackage{booktabs}
\usepackage{algorithm}
\usepackage{algorithmic}
\usepackage{subfiles}
\usepackage{amssymb}
\usepackage[switch]{lineno}
\usepackage{multirow}
\usepackage{multicol}
\usepackage[table,xcdraw]{xcolor}
\usepackage{efbox}
\usepackage{boldline} 
\usepackage{tabu}

\DeclareMathOperator*{\argmin}{arg\,min}

\title{A Survey of Supernet Optimization and its Applications:\\
Spatial and Temporal Optimization for Neural Architecture Search}

\author{
    Stephen Cha, Taehyeon Kim, Hayeon Lee, Se-Young Yun
    \affiliations
    KAIST AI
    \emails
    \{jooncha, potter32, hayeon926, yunseyoung\}@kaist.ac.kr,
}

\begin{document}

\maketitle

\begin{abstract}

This survey focuses on categorizing and evaluating the methods of supernet optimization in the field of Neural Architecture Search (NAS). Supernet optimization involves training a single, over-parameterized network that encompasses the search space of all possible network architectures. The survey analyses supernet optimization methods based on their approaches to spatial and temporal optimization. Spatial optimization relates to optimizing the architecture and parameters of the supernet and its subnets, while temporal optimization deals with improving the efficiency of selecting architectures from the supernet. The benefits, limitations, and potential applications of these methods in various tasks and settings, including transferability, domain generalization, and Transformer models, are also discussed.
\end{abstract}

\section{Introduction}\label{sec1}
Designing neural networks is a crucial part of deep learning, but it requires a lot of time and effort from experts without any guarantee that the resulting architectures will be transferable to different tasks or settings. To address this challenge, Neural Architecture Search (NAS) automates the design process by effectively sampling viable models from a large space of potential architectures. NAS has received a lot of attention from both academia and industry due to its potential applications in various contexts, such as semantic segmentation in satellite imagery \cite{gudzius2022automl}, environments with limited computational resources like the Internet of Things (IoT) \cite{zeinali2021esai}, and user-friendly AutoML packages.

Supernet optimization is a method within Neural Architecture Search (NAS) that trains a single, highly parameterized neural network, known as the supernet, to represent the search space of all possible architectures. This is done through weight-sharing, where subnets can inherit parameters from the supernet and then feed their trained parameters back into it. This process repeats, resulting in subnets that continually inherit weights from the updated supernet. This approach offers several benefits, including the ability to transfer architectures across different tasks, hardware, and dataset domains, resulting in improved energy efficiency \cite{cai2019once,zoph2018learning}. The most significant advantage of supernet optimization is that it allows for multiple deployable architectures to be generated from a single training of the supernet, meaning ``train once, sample many".

This survey endeavors to classify and examine supernet optimization techniques according to their methods of \textbf{spatial} and \textbf{temporal} optimizations. The former refers to optimizing the architecture and parameters of the supernet and its derived subnets, while the latter pertains to enhancing the ease and speed of selecting architectures from the supernet. The survey will also highlight the advantages and limitations of these methods and their potential uses in transferring to different tasks and settings, domain generalization, and Transformer models. We also provide a table that groups methods with respect to our proposed taxonomy (Table \ref{tab:overview}). 

\begin{figure*}[!ht]
    \centering
    \hbox{\hspace{-0.5cm}\includegraphics[width=\textwidth,keepaspectratio]{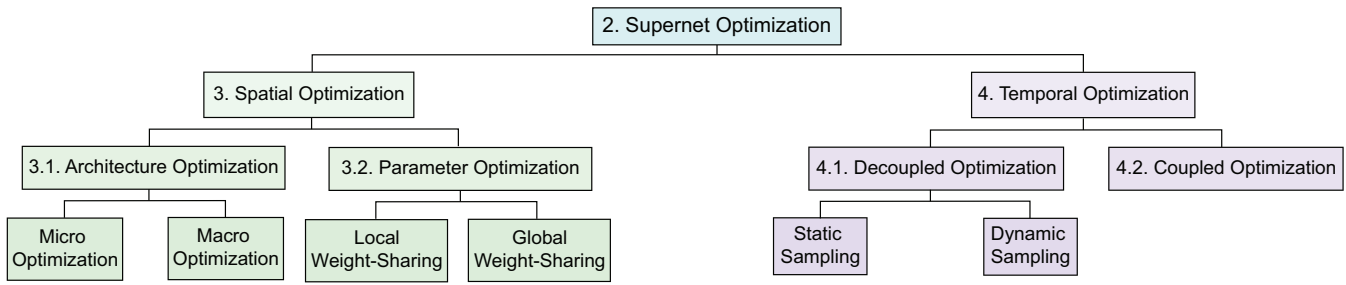}}
    \caption{Supernet optimization taxonomy with respect to spatial and temporal optimizations. }
    \label{fig:overview}
\end{figure*}

\section{Supernet Optimization}
\subsection{Neural Architecture Search}
Neural Architecture Search (NAS) is a technique for automatically finding the best architecture for a deep learning task. It is designed to balance performance and resource constraints. NAS is made up of three key components: the search space, the search strategy, and the performance estimation strategy \cite{elsken2019neural}.

\begin{itemize}
  \item \textbf{Search space:} The search space defines all possible architectures for the task at hand. It is important to note that introducing inductive bias into the search space is inevitable. One of the earliest forms of a search space was a simple sequential layer of operations.
  \item \textbf{Search strategy:} The search strategy determines how the architectures are selected from the search space. There are two main objectives: finding high-performing architectures and minimizing search time and memory footprint. One approach to balance these goals is to explore the Pareto frontier, which is the set of solutions where improving one objective will cause another objective to worsen.
  \item \textbf{Performance estimation strategy:} The performance estimation strategy measures how well the selected architecture performs on new unseen data. The most straightforward approach is to predict performance on the validation set, but this is computationally expensive. Finding a fast and effective proxy metric is still an open question in the NAS field.
\end{itemize}

\subsection{Supernet: Train Once, Sample Many}\label{supernet}
A \textbf{supernet} is a neural network that represents the search space, including all possible architectures to be selected. It can be visualized as a Directed Acyclic Graph (DAG), where nodes are feature maps or module blocks and edges are operations (e.g., convolution, pooling, identity, zero).

Subnets are derived from the supernet and trained, with their parameters then combined back into the supernet. This process is repeated and each subsequent subnet inherits weights from the supernet, which holds the parameters from previously trained subnets - this is referred to as \textbf{weight-sharing} \cite{pham2018efficient}. This approach has made the network search process more efficient, reducing the time required from thousands of GPU days to just a few GPU days, compared to non-weight-sharing methods \cite{zoph2016neural,real2017large}.

While weight-sharing has improved efficiency, it suffers from a poor correlation between the supernet optimization and the standalone performances. This survey will look into methods to address this correlation problem through a combination of spatial and temporal optimization. Additionally, supernet optimization allows for direct deployment after optimization \cite{cai2019once,yu2020bignas}, which means that it only needs to be trained once to generate many different network architectures. The survey will also cover various applications of supernet optimization, including the transfer of network architectures to different tasks, domain generalization, and Transformer models.

\section{Spatial Optimization}
The effectiveness of a neural network in learning from data is based on both its architecture and its parameters \cite{li2020block}. Spatial optimization involves optimizing both the architecture and parameters of the supernet and its subnets as explained in Section \ref{supernet}. The supernet is typically shown as a Directed Acyclic Graph (DAG), where the nodes are feature maps or blocks and the edges describe the operations performed. The parameters of the network are stored in the DAG.

To find an effective and efficient architecture, the search space needs to have the capacity to contain it. The supernet serves as the search space, so careful construction is crucial. The supernet can be parameterized by various elements, such as a list of possible operations, the number of nodes or layers, layer widths, edge connectivity, and other configurations specific to the model (e.g. kernel sizes for convolutional architectures). In this survey, the process of predefining or optimizing these configurations is referred to as architecture optimization (Section~\ref{subsec:arch_opt}).

The issue with supernet optimization is poor rank correlation, leading to subnets often needing to be retrained from scratch, which is inefficient and limits the ``train once, sample many" property. To address this, methods that perform supernet optimization in a way that subnets can be directly deployed or need only minor fine-tuning are referred to as parameter optimization (Section~\ref{subsec:param_opt}).

\subsection{Architecture Optimization}\label{subsec:arch_opt}
Designing a rich search space of viable neural network architectures for various tasks is critical for efficiently and effectively discovering the highest-performing architecture. The modern view of neural networks sees them as a graph of blocks, such as residual or MBConv blocks, where each block is also represented as a graph of feature maps connected by operations like convolution, pooling, or identity. The process of creating blocks from scratch is known as \textbf{micro optimization}, while \textbf{macro optimization} involves finding the optimal combination of blocks, either predefined or created through micro optimization, to form a neural architecture.

\subsubsection{Micro Optimization}

Micro optimization \cite{liu2018darts,pham2018efficient} refers to constructing a single block (often referred to as a cell) from scratch by selecting from a predefined set of operations. This block is then repeated across the network, resulting in a homogeneous architecture. This concept was inspired by NASNet \cite{zoph2018learning}, which was influenced by the repeated block structures commonly found in successful neural networks. Micro design incurs the burden of the discovered cell needing high representation power when repeating itself across the network structure. While micro optimization has been effective in practice, there is a lack of theoretical understanding as to why this is the case.

In addition to the combination of operations in a block, the connectivity within the block also plays a crucial role in determining the model's capacity. One popular heuristic for defining the connectivity is to predefine the topology such that each intermediate node takes input from the two operations with the highest magnitudes \cite{liu2018darts}. However, this heuristic lacks theoretical and experimental evidence to support its use. To address this issue, model pruning approaches have been proposed, such as introducing a differentiable architecture parameter on the edges \cite{yang2021towards} or decoupling the topology and operation search \cite{gu2021dots}. Sparsity can also be employed to remove handcrafted heuristics and yield a more flexible cell \cite{wu2021neural}.

\subsubsection{Macro Optimization}

Macro optimization is a paradigm that aims to determine an optimal combination of blocks, resulting in a heterogeneous network architecture \cite{cai2019once,wu2021fbnetv5,guo2020single}. The blocks are usually predefined, either by handcrafted designs or by micro optimization. To reduce inductive bias from human-designed blocks, micro optimization can be used to search for candidate blocks, which can then be combined with macro optimization.

Based on predefining a set of blocks, one can abstract the supernet as if it is composed of elementary ``atoms". \cite{mei2019atomnas} introduces atomic blocks, which are composed of two convolutions and a channel-wise operation, which are used to define a fine-grained search space. The supernet can be pruned by penalizing blocks with a high FLOP count. Another pruning-based approach is to remove operations that are rarely or never sampled during supernet optimization \cite{xia2022progressive}, which reduces unnecessary computational overhead wasted on superfluous operations. Data-based network pruning can also be applied \cite{dai2020data}, where blocks are pruned out if they underperform on an easier task with fewer data labels.

In macro optimization, inter-block connectivity is also crucial. The naive approach is to sequentially stack blocks with no skip connections, however, dense connectivity \cite{fang2020densely} can be used to form the search space. The dense connectivity between blocks is then determined by a differentiable search.


\subsection{Parameter Optimization}\label{subsec:param_opt}

Supernet optimization offers the advantage of allowing subnets to inherit parameters from the supernet during training, thus reducing computational overhead. However, naive parameter optimization results in subnets with poor rank correlation. A random search baseline, in which subnets are randomly sampled and retrained from scratch, has been shown to outperform subnets sampled from the supernet with inherited weights \cite{gong2021nasvit}. This highlights the need for a more sophisticated approach to training supernet parameters to ensure that directly deployed subnets perform well. To achieve this goal, the weight-sharing mechanism must be optimized. Two approaches include \textbf{local weight-sharing}, where the subnet weights are shared across a subregion of the supernet as opposed to the entire space, and \textbf{global weight-sharing} which improves the vanilla weight-sharing method without requiring a sub-supernet.
 
\subsubsection{Local Weight-Sharing} \label{sssec:local}

The optimization of a supernet typically allows subnets to inherit parameters from the supernet during training. However, naively optimizing parameters leads to subnets with poor rank correlation. A random search baseline that trains subnets from scratch outperforms subnets inherited from the supernet \cite{gong2021nasvit}. Thus, there is a need to develop an intelligent training mechanism for supernet parameters to ensure proper performance upon deployment. One approach is to use local weight-sharing, which can be applied to a subset of the supernet, or global weight-sharing, which improves traditional weight-sharing without reducing the size of the supernet \cite{gong2021nasvit}.

Vanilla weight-sharing across the entire supernet leads to instability in optimization \cite{bender2018understanding}. Local weight-sharing across a subset of the supernet improves rank correlation \cite{li2020block,chu2021fairnas,zhao2021few}. The supernet can be partitioned into blocks for fair subnet training \cite{chu2021fairnas}. One method involves training the supernet block-wise by optimizing the block-wise supernet weights and architecture parameters \cite{li2020block,moons2021distilling,li2021bossnas}. Another approach is to partition the search space hierarchically by splitting the connections within the supernet. The hierarchy is established by training the whole supernet and then partitioning it based on a random edge. Sub-supernets inherit parameters from their parent supernets for initialization, then the sub-supernets are trained to convergence. Upon architecture search, the subnet will inherit parameters from the corresponding sub-supernet \cite{zhao2021few}.

Entangling weights across different choice modules can also improve supernet optimization \cite{chen2021autoformer}. This method updates weights simultaneously across all choice modules, causing the weights of each choice module to be entangled. To close the distribution gap between supernet weights and standalone weights, a regularization term is introduced \cite{pan2022distribution}. During each iteration of supernet optimization, two architectures have been sampled that share at least one operator. The parameters of the current subnet are updated with gradients from both the current and other subnets. The learning rate for the other subnet is determined by a meta-network that measures the degree of matching between the two architectures.

Finally, a pruning-based local weight-sharing technique can also be used for parameter optimization \cite{cai2019once}. The supernet is trained by taking progressively smaller subnets with respect to depth, width, kernel size, and input resolution. Although this technique is designed for convolutional architectures, it can be applied to any architecture type.

\subsubsection{Global Weight-Sharing}
Global weight-sharing refers to methods that weight-share across the entire supernet aiming to alleviate poor rank correlation. Due to little theoretical understanding of the direct relationship between weight-sharing and poor rank correlation, it is unclear whether a local or global weight-sharing technique is universally better than the other. Regardless of the truth, there is abundant work that integrates both local and global weight-sharing techniques to achieve performance \cite{li2020block,moons2021distilling,li2021bossnas}. 

The most commonly used weight-sharing technique in global weight-sharing is In-place Knowledge Distillation (KD) \cite{cai2019once,yu2020bignas}. This method trains the supernet using ground-truth labels, while subnets are trained using soft labels generated by the supernet. The weights of the supernet are updated by aggregating the gradients of the subnets. KL divergence is a common metric used to minimize the difference between the teacher and student distributions. However, it was shown in \cite{wang2021alphanet} that using KL divergence can lead to the student subnets overestimating or underestimating the uncertainty of the teacher model. An alternative KD loss function was proposed to address this issue.

One concern with KD-based supernet optimization is the reliance on a pretrained teacher model, which can create an inductive bias towards handcrafted architectures \cite{li2020block,moons2021distilling}. To avoid this, some works have employed self-supervised learning using siamese supernets \cite{li2021bossnas}. Another approach is to use a board of prioritized paths in the supernet to serve as teachers for undertrained subnets \cite{peng2020cream}.

Another issue with weight-sharing is the risk of biasing updates towards a specific subnet, causing the supernet to forget previous knowledge and leading to poor rank correlation \cite{chu2021fairnas}. To address this, \cite{ha2022sumnas} proposed a meta-learning approach where the supernet's parameters learn meta-features for multiple subnets during optimization. This allows the supernet to use its learned parameters for direct deployment across different subnets.

\section{Temporal Optimization}
In order to accommodate many different models that achieve high performance across various tasks, it is necessary to construct a large search space rich with powerful neural architectures. Supernets tend to have an exponentially large amount of subnets. In order to do proper weight-sharing across the subnets, it is imperative to address the tractability and efficiency of sampling such an architecture. We refer to such optimization as \textbf{temporal optimization}. 

Supernet optimization in one-shot NAS \cite{bender2018understanding,cai2019once} samples a subnet and trains, then the subnet parameters are weight-shared back to the supernet, referring to such process as \textbf{decoupled optimization}. Otherwise, differentiable NAS \cite{liu2018darts,mei2019atomnas} the sampling process is relaxed to a continuous parameter which is used to solve a bilevel optimization between the architecture and its network parameters. We refer to the such process as \textbf{coupled optimization}. Note that we refer to coupled optimization and differentiable search interchangeably. 

\subsection{Decoupled Optimization}
Let $\mathcal{A}$ be the search space and let $\mathcal{A}_{good}$ be a subset of ``good" architectures such that $\mathcal{A}_{good}\subset \mathcal{A}$. The precise construction of $\mathcal{A}_{good}$ is dependent on the needs of the user. For example, the set of good architectures can be all architectures that achieve a certain performance metric given a certain resource constraint. The sampling distribution need not realize every single $a\in\mathcal{A}$, but it does need to strive to realize every $a\in\mathcal{A}_{good}$. The goal is to assign probability masses across the search space such that the chances of a ``good" architecture being sampled are higher. This will lead to various better subnets being properly trained during supernet optimization, which will lead to better performances. 

The distribution of probability masses across the search space can be categorized by: \textbf{static} and \textbf{dynamic} distributions. We note that there is a trade-off between the two distribution types. Static distributions require the user to make a choice on the distribution of architectures, which might be naive depending on the search space and task. Since the probabilities are determined, there is no need to compute them during the optimization process. In contrast, setting the distribution to be dynamic requires real-time additional computations during supernet optimization. However, a dynamic distribution is more likely to focus training on viable architecture than a naive fixed prior. 

\subsubsection{Static Distribution}

In \cite{guo2020single}, the authors view the supernet as a set of choice modules (macro design) and randomly select a single subnet for training during the supernet optimization phase. However, this approach can lead to biased supernets where the parameters of certain choice modules are updated more frequently than others. To address this, \cite{chu2021fairnas} proposes a method to ensure that the parameters of each choice module are updated the same number of times. The approach described in \cite{guo2020single} is referred to as single path sampling, which only samples a single architecture (subnet) for optimization. Multi-path sampling is an extension to this method, where multiple architectures are sampled for supernet optimization. In \cite{chu2020mixpath}, the authors find that mixing feature vectors from multiple paths can cause instability during training. They resolve this issue by scaling down the feature vectors to have the same magnitude.

\subsubsection{Dynamic Distribution}

Search spaces in NAS are vastly large. A supernet cell with $m$ layers and $n$ candidate operations would have a total of $n^m$ subnets. Uniform sampling during supernet optimization on such a vast search space might require larger training costs due to wasted compute time on ``bad" architectures. The intuitive approach to dynamically reshaping the sampling distribution is to apply a greedy approach. \cite{you2020greedynas} present a method that exploits good architectures from a candidate pool and explores the search space by adding new subnets to the pool.  The authors also introduce an exploration procedure by adding uniformly sampled subnets from the search space to the candidate pool. Another greedy approach \cite{li2020sgas} uses model pruning to remove ineffective subnets. Another approach, called attentive sampling, is to find the Pareto front of the search space and sample $k$ subnets, training the one with the best or worst predicted performance \cite{wang2021attentivenas}. The aim of training the subnet with the worst predicted performance is to make use of it as a difficult example and improve exploration in the search space.

A recent discovery \cite{hu2022generalizing,gong2021nasvit} suggests that addressing gradient conflicts can mitigate poor rank correlation. During supernet optimization, the gradient updates from separate subnets can contradict each other and delay or prevent convergence. \cite{gong2021nasvit} alleviates conflicting gradients by projecting supernet gradients to the normal vector of the subnet gradients which removes the conflicting components. In addition, they propose switchable channel-wise scaling layers for each Transformer layer. The aim is to allow subnets with different layer widths and depths to rescale their features in a local way. In addition, they observed that gradient conflicts are more prone to happen in the presence of strong data augmentation and regularization. \cite{hu2022generalizing} improves upon few-shot NAS \cite{zhao2021few} by replacing uniform sampling in the hierarchical partitioning with gradient matching, i.e., edges with operations that yield contradicting gradients will be given priority in the split. This keeps operations with conflicting gradients in separate sub-supernets and alleviates the unstable optimization. While not directly aligning gradients, \cite{xu2022analyzing} introduce a local sampling method that samples a subnet such that it differs by only one operation from the previous subnet. 

\subsection{Coupled Optimization}
The differentiable search \cite{liu2018darts} gives each edge in the DAG as an architecture parameter weight $\alpha$. Differentiable search is distinct from decoupled optimization because it does not actually sample a subnet during the training phase, but rather trains each subnet in a nested manner, i.e., solves a bilevel optimization problem to determine the searched architecture: 
\begin{equation}
    \begin{aligned}
        &\alpha^*=\argmin_{\alpha\in\mathcal{A}}{\mathcal{L}_{val}(\alpha,w^*(\alpha))}\\
        \text{s.t. }&w^*(\alpha)=\argmin_{w}{\mathcal{L}_{train}(\alpha,w)}
    \end{aligned}
\end{equation}
where $w$ is the supernet parameters with architecture topology configuration $\alpha$, $\mathcal{A}$ denotes the search space, and $\mathcal{L}$ is the loss function. Differentiable search has garnered popularity due to reducing computation time from thousands of GPU days to a few GPU hours and allowing architecture search to run on a single GPU. However, the early works suffered from various issues: optimization instability, memory overhead, and poor generalizability. However, coupled optimization is still a valuable avenue of supernet research due to the automated grouping of supernet training and searching. 

Differentiable architecture search yields a cell module and measures performance after stacking eight cells vertically. After choosing the cell that yields the highest accuracy, the same cell is stacked twenty times for final evaluation. This separated process results in independent optimization during the search and evaluation process. Arguing that this separate optimization process causes instability, \cite{yu2022cyclic} transfer model parameters and architecture knowledge from the evaluation network to the search network to improve stability. Other issues are alleviated by \cite{xu2019pc} improving memory overhead by introducing partial-channel sampling during the training phase, \cite{zela2019understanding} improving generalization by applying loss function sharpness aware regularization, and \cite{wang2021rethinking} improving generalization by retaining the differentiable training process and simply introducing a perturbation-based architecture selection method that discretizes the supernet by taking a random edge and checking its performance accuracy as opposed to taking the largest architecture weight value, which demonstrated that the architecture parameter might not be actually representative of how viable an operation is. Also, \cite{shu2019understanding} demonstrate empirically and theoretically that differentiable search converges to wider and shallower architectures due to smoother loss landscapes and smaller gradient variance compared to a random search baseline despite poor generalization. A way to circumvent biasing towards wide and shallow networks is to remove the top-2 input connection heuristic in the connectivity \cite{gu2021dots}.

\efboxsetup{linecolor=blue!75!black, linewidth=2pt, margin=0pt}
\begin{table*}[htp!]
\centering
\caption{Table of supernet-based methods that are mentioned in the survey. The works are categorized with respect to the proposed spatial and temporal optimization framework along with applications.\label{tab:overview}}
\renewcommand*{\arraystretch}{1.5}
\resizebox{\textwidth}{!}{%

\begin{tabular}{|
>{\columncolor[HTML]{E2FFE8}}c 
>{\columncolor[HTML]{E2FFE8}}c 
>{\columncolor[HTML]{E2FFE8}}c |l|}
\hline
\multicolumn{1}{|c|}{\cellcolor[HTML]{E2FFE8}}                                                                                               & \multicolumn{1}{c|}{\cellcolor[HTML]{E2FFE8}}                                                                                                     & \textbf{\begin{tabular}[c]{@{}c@{}}Micro \\ Optimization \end{tabular}}                      & \begin{tabular}[l]{@{}l@{}}DOTS [\citeyear{gu2021dots}], SGAS [\citeyear{li2020sgas}], DARTS [\citeyear{liu2018darts}], ENAS [\citeyear{pham2018efficient}], DARTS+PT [\citeyear{wang2021rethinking}], SparseNAS [\citeyear{wu2021neural}],\\ EnTranNAS\,[\citeyear{yang2021towards}], CDARTS\,[\citeyear{yu2022cyclic}]\end{tabular} \\ \cline{3-4} 
\multicolumn{1}{|c|}{\cellcolor[HTML]{E2FFE8}}                                                                                               & \multicolumn{1}{c|}{\multirow{-2}{*}[0.9em]{\cellcolor[HTML]{E2FFE8}\textbf{\begin{tabular}[c]{@{}c@{}}3.1. Architecture \\ Optimization\end{tabular}}}} & \textbf{\begin{tabular}[c]{@{}c@{}}Macro \\ Optimization\end{tabular}}                      & \begin{tabular}[l]{@{}l@{}}DA-NAS\,[\citeyear{dai2020data}], HR-NAS\,[\citeyear{ding2021hr}], DenseNAS\,[\citeyear{fang2020densely}], DNA\,[\citeyear{li2020block}], BossNAS\,[\citeyear{li2021bossnas}], AtomNAS\,[\citeyear{mei2019atomnas}],\\
DONNA\,[\citeyear{moons2021distilling}], PAD-NAS\,[\citeyear{xia2022progressive}]\end{tabular} \\ \cline{2-4} 
\multicolumn{1}{|c|}{\cellcolor[HTML]{E2FFE8}}                                                                                               & \multicolumn{1}{c|}{\cellcolor[HTML]{E2FFE8}}                                                                                                     & \textbf{\begin{tabular}[c]{@{}c@{}}Local \\ Weight-Sharing\end{tabular}}                    & \begin{tabular}[l]{@{}l@{}}OFA\,[\citeyear{cai2019once}], AutoFormer\,[\citeyear{chen2021autoformer}], GM-NAS\,[\citeyear{hu2022generalizing}], DNA\,[\citeyear{li2020block}], BossNAS\,[\citeyear{li2021bossnas}], ViTAS\,[\citeyear{su2022vitas}],\\
BigNAS\,[\citeyear{yu2020bignas}], Few-Shot-NAS\,[\citeyear{zhao2021few}]\end{tabular}  \\ \cline{3-4} 
\multicolumn{1}{|c|}{\multirow{-4}{*}[2.25em]{\cellcolor[HTML]{E2FFE8}\textbf{\begin{tabular}[c]{@{}c@{}}3. Spatial \\ Optimization\end{tabular}}}}  & \multicolumn{1}{c|}{\multirow{-2}{*}[0.9em]{\cellcolor[HTML]{E2FFE8}\textbf{\begin{tabular}[c]{@{}c@{}}3.2. Parameter \\ Optimization\end{tabular}}}}    & \textbf{\begin{tabular}[c]{@{}c@{}}Global \\ Weight-Sharing\end{tabular}}                   & \begin{tabular}[l]{@{}l@{}}OFA\,[\citeyear{cai2019once}], DA-NAS\,[\citeyear{chang2022data}], AutoFormer\,[\citeyear{chen2021autoformer}],MixPath\,[\citeyear{chu2020mixpath}], FairNAS\,[\citeyear{chu2021fairnas}], HR-NAS\,[\citeyear{ding2021hr}],\\
SUM-NAS\,[\citeyear{ha2022sumnas}], DNA\,[\citeyear{li2020block}], BossNAS\,[\citeyear{li2021bossnas}], Cream\,[\citeyear{peng2020cream}], AlphaNet\,[\citeyear{wang2021alphanet}], BigNAS\,[\citeyear{yu2020bignas}]\end{tabular} \\ \hline
\multicolumn{1}{|c|}{\cellcolor[HTML]{E7E1F5}}                                                                                               & \multicolumn{1}{c|}{\cellcolor[HTML]{E7E1F5}}                                                                                                     & \cellcolor[HTML]{E7E1F5}\textbf{\begin{tabular}[c]{@{}c@{}}Static \\ Sampling\end{tabular}} & \begin{tabular}[l]{@{}l@{}}One-Shot\,[\citeyear{bender2018understanding}], SPOS\,[\citeyear{guo2020single}], OFA\,[\citeyear{cai2019once}], DATA\,[\citeyear{chang2022data}], SUM-NAS\,[\citeyear{ha2022sumnas}], DNA\,[\citeyear{li2020block}],\\
AtomNAS\,[\citeyear{mei2019atomnas}], DONNA\,[\citeyear{moons2021distilling}], HAT\,[\citeyear{wang2020hat}], AlphaNet\,[\citeyear{wang2021alphanet}], BigNAS\,[\citeyear{yu2020bignas}]\end{tabular}   \\ \cline{3-4} 
\multicolumn{1}{|c|}{\cellcolor[HTML]{E7E1F5}}                                                                                               & \multicolumn{1}{c|}{\multirow{-2}{*}[0.85em]{\cellcolor[HTML]{E7E1F5}\textbf{\begin{tabular}[c]{@{}c@{}}4.1. Decoupled \\ Optimization\end{tabular}}}}    & \cellcolor[HTML]{E7E1F5}\textbf{\begin{tabular}[c]{@{}c@{}}Dynamic\\ Sampling\end{tabular}} &  \begin{tabular}[l]{@{}l@{}}NAS-ViT\,[\citeyear{gong2021nasvit}], GM-NAS\,[\citeyear{hu2022generalizing}], SGAS\,[\citeyear{li2020sgas}], DC-NAS\,[\citeyear{pan2022distribution}], ENAS\,[\citeyear{pham2018efficient}], AttentiveNAS\,[\citeyear{wang2021attentivenas}]\\
MAGIC\,[\citeyear{xu2022analyzing}], GreedyNAS\,[\citeyear{you2020greedynas}], Few-Shot-NAS\,[\citeyear{zhao2021few}]\end{tabular} \\ \cline{2-4} 
\multicolumn{1}{|c|}{\multirow{-3}{*}[1.6em]{\cellcolor[HTML]{E7E1F5}\textbf{\begin{tabular}[c]{@{}c@{}}4. Temporal \\ Optimization\end{tabular}}}} & \multicolumn{2}{c|}{\cellcolor[HTML]{E7E1F5}\textbf{\begin{tabular}[c]{@{}c@{}}4.2. Coupled \\ Optimization\end{tabular}}}                                                                                                                      & \begin{tabular}[l]{@{}l@{}}NAS-OOD\,[\citeyear{bai2021ood}], DenseNAS\,[\citeyear{fang2020densely}], DOTS\,[\citeyear{gu2021dots}],SGAS\,[\citeyear{li2020sgas}], DARTS\,[\citeyear{liu2018darts}], DART+PT\,[\citeyear{wang2021rethinking}],\\
SparseNAS\,[\citeyear{wu2021neural}], PAD-NAS\,[\citeyear{xia2022progressive}], PC-DARTS\,[\citeyear{xu2019pc}], EnTranNAS\,[\citeyear{yang2021towards}], CDARTS\,[\citeyear{yu2022cyclic}], RDARTS\,[\citeyear{zela2019understanding}]\end{tabular}  \\ \hline
\multicolumn{3}{|c|}{\cellcolor[HTML]{FFD3D3}\textbf{5. Extensions to Applications}}      & \begin{tabular}[l]{@{}l@{}}OFA\,[\citeyear{cai2019once}], BigNAS\,[\citeyear{yu2020bignas}], DONNA\,[\citeyear{moons2021distilling}], DATA\,[\citeyear{chang2022data}], FBNetV5\,[\citeyear{wu2021fbnetv5}], SPIDER\,[\citeyear{mushtaq2021spider}], FedSup\,[\citeyear{kim2022supernet}],\\ NAS-DA\,[\citeyear{li2020network}], NAS-OOD\,[\citeyear{bai2021ood}], AutoFormer\,[\citeyear{chen2021autoformer}], ViTAS\,[\citeyear{su2022vitas}], NAS-ViT\,[\citeyear{gong2021nasvit}], HR-NAS\,[\citeyear{ding2021hr}], \\HAT\,[\citeyear{wang2020hat}], MAGIC\,[\citeyear{xu2022analyzing}] \end{tabular}\\ \hline
\end{tabular}

}

\end{table*}

\section{Extensions to Applications}











Supernets have become the state-of-the-art method in the field of neural architecture search (NAS) and have various relevant applications outside of the NAS research area.

\subsection{Transferable Neural Architectures}
\subsubsection{Specialized Networks for Specific Hardware Settings}
Deep learning is performed on various hardware platforms, ranging from powerful high-performance GPU clusters to small mobile devices. Supernets offer the advantage of "train once, sample many", where the supernet can be trained such that its subnets can be deployed to different hardware platforms with specific constraints, such as FLOPs, memory size, and run-time latency. For example, local weight-sharing by iterative pruning with respect to the number of layers, channels, kernel sizes, and input resolution can be used to train the supernet and make the subnets directly deployable to a mobile device \cite{cai2019once}. The global weight-sharing through inplace distillation by making the smaller subnets learn from the supernet can also be used \cite{cai2019once,yu2020bignas}. Hardware-aware search across diverse search spaces has also been generalized \cite{moons2021distilling}.

\subsubsection{Tailored Models for Different Datasets \& Applications }

NAS methods are usually optimized for a single task, but supernets have the capacity to perform well on various tasks and datasets, making them suitable for self-supervised approaches. For example, a self-supervised approach can be used where the largest subnet acts as a teacher model, and two student subnets are used \cite{chang2022data}. Another approach involves aggregating loss functions for each task by computing a weighted sum of the task losses \cite{wu2021fbnetv5}.

Another avenue of transferability would be tasks such as different datasets\,(e.g., CIFAR-10 vs ImageNet) and learning applications\,(e.g., image classification, semantic segmentation, language modeling). Most NAS methods are optimized with only one task considered. This would require the supernet optimization to be repeated for each individual task, which would incur high computational costs. Given that a supernet is large and expected to have the capacity to perform well on a number of different tasks, a self-supervised approach can be done \cite{chang2022data} to achieve task transferability. The supernet optimization is done by a self-supervised approach by setting the largest subnet as the teacher model, sampling two student subnets, then applying contrastive learning between the teacher and student models. Another approach is supernet optimization by aggregating loss functions for each task \cite{wu2021fbnetv5}. The aggregation is done by computing a weighted sum of the $T$ task losses where the weights are collected by importance sampling. 

\subsubsection{Supernets for Federated Learning}
Supernets are transferable and thus suitable for the federated learning (FL) scenario where neural networks are deployed across various devices and resource constraints. A straightforward approach is to give each client a supernet and apply local device training to search for personalized architectures \cite{mushtaq2021spider}. Another approach is to give a client a subnet from the supernet for optimization on the local device \cite{kim2022supernet}.

\subsection{Supernet Domain Generalization}
NAS research mostly focuses on clean datasets, but real-world settings often lack clean data and have distribution discrepancies between training and testing stages. To address these issues, supernet optimization needs to be robust to Out-of-Distribution (OOD) cases. Adversarial learning can be applied to supernet optimization to make the searched architectures robust to distribution mismatches. For example, a two-stage method can be used where a regularization term is introduced to measure the discrepancy between source and target domains, followed by adversarial learning \cite{li2020network}. Differentiable NAS with an adversarial minimax optimization problem can also be used \cite{bai2021ood}.

\subsection{Transformers Meet Supernets}
The widespread use and exceptional performance of Transformer architectures make it deserving of its own subsection. Selecting the optimal depth, embedding dimension, and number of heads for a Transformer model is a specialized task that requires significant knowledge, making it challenging for practitioners to achieve state-of-the-art performance. The state-of-the-art Transformer models for vision tasks require billions of parameters and thousands of GPUs, but supernet optimization can achieve comparable results with a reduced parameter count of millions. This also reduces the need for fine-tuning and makes it easier for practitioners to build and train Transformer models for specific tasks using limited computing resources.

Supernet optimization has resulted in Transformer networks that achieve performance comparable to handcrafted models but with increased efficiency (reduced FLOPs and parameter count). However, straightforward supernet optimization on Transformer models has a tendency to result in slow convergence and low performance. To address this, a local weight-sharing technique, which interconnects the weights of similar Transformer modules (refer to Section \ref{sssec:local}), has been proposed to allow for faster convergence, reduced memory usage, and improved subnet performance \cite{chen2021autoformer}. Another local weight-sharing method was proposed in \cite{su2022vitas} to enhance the fairness and uniformity of training through a cyclic weight-sharing mechanism across the channels. \cite{gong2021nasvit} resolved conflicting gradient updates in supernet optimization through gradient projection and introduced switchable channel-wise scaling layers for each Transformer layer. Another development is supernet optimization for lightweight Transformer models specifically designed for high-resolution image tasks \cite{ding2021hr}. In addition to vision tasks, there are also supernet optimization methods for Transformer-based models in NLP tasks \cite{wang2020hat,xu2022analyzing} that take into consideration hardware constraints and are temporally optimized.

\section{Future Directions}
A major challenge in supernet optimization is the lack of stability in the optimization process, commonly referred to as poor rank correlation. Many studies have tried to tackle this problem using spatial or temporal optimization techniques, but the underlying cause of the issue remains unclear. Previously, it was thought that weight sharing had an inherent caveat where passing on parameters across different subnets was inappropriate for proper performance. However, recent research suggests that conflicting gradient updates during optimization may also play a role \cite{hu2022generalizing,gong2021nasvit}. 

Although temporal optimization has been widely used to find efficient architectures, the use of meta-learning-based embedding spaces and performance predictors has received little attention in the supernet optimization literature, cite{lee2021rapid,lee2021hardware}. This approach allows for efficient approximation of architecture performance and incorporating weight sharing with latent space optimization and meta-learned performance predictors could further enhance supernet transferability.

At its essence, the goal of supernet optimization is to find the best neural architecture. To achieve this, a rich search space of models and an efficient sampling algorithm must be designed. Although human-designed biases have produced good results, it would be beneficial to reduce human intervention in the optimization process. The recent success of evolutionary search in reconstructing the backpropagation algorithm from basic mathematical operations \cite{real2020automl} suggests that minimizing human biases in supernet optimization could lead to further advancements.

\section{Conclusion}
Supernet optimization is a powerful tool for finding effective and efficient neural network architectures for different settings and tasks (as shown in Table \ref{tab:overview}). However, a straightforward approach can produce models that perform worse than a random search baseline, according to studies like \cite{shu2019understanding} and \cite{gong2021nasvit}. Spatial optimization helps to find models that are both powerful and efficient by optimizing both the architecture and parameters of a network. Temporal optimization enables efficient access to these powerful models and ensures fair training. Supernet models are accessible from a search perspective and are ready to be deployed after optimization, making them suitable for transferring across hardware settings. In addition, they have been demonstrated to be useful in various applications, including domain generalization and Transformer models.

\appendix



\bibliographystyle{named}
\bibliography{main}

\end{document}